\documentclass{article}
\usepackage[english]{babel}
\usepackage{inputenc}
\usepackage{csquotes}
\usepackage{fancyhdr}
\usepackage{subfigure}
\usepackage[final,nonatbib]{nips_2018}
    \usepackage{adjustbox}
    \usepackage{enumitem}
    \usepackage{enumerate}
\usepackage{color}
\usepackage{booktabs}      
\usepackage{nicefrac}       
\usepackage{graphicx}
\graphicspath{ {./images/} }
\title{Split learning for health: Distributed deep learning without sharing raw patient data}
\author{
  Praneeth Vepakomma\thanks{Corresponding author, e-mail: vepakom@mit.edu} \\
  Massachusetts Institute of Technology\\
  Cambridge, MA 02139 \\
   \And
   Otkrist Gupta \\
   Massachusetts Institute of Technology \\
   Cambridge, MA 02139 \\
   \And
   Tristan Swedish \\
   Massachusetts Institute of Technology \\
   Cambridge, MA 02139 \\
   \And
   Ramesh Raskar \\
   Massachusetts Institute of Technology \\
   Cambridge, MA 02139 \\
}

\begin{document}
\maketitle
\fontsize{10}{18}

\begin{abstract}

Can health entities collaboratively train deep learning models without sharing sensitive raw data? This paper proposes several configurations of a distributed deep learning method called SplitNN to facilitate such collaborations. SplitNN does not share raw data or model details with collaborating institutions. The proposed configurations of splitNN cater to practical settings of i) entities holding different modalities of patient data, ii) centralized and local health entities collaborating on multiple tasks and iii) learning without sharing labels. We compare performance and resource efficiency trade-offs of splitNN and other distributed deep learning methods like federated learning, large batch synchronous stochastic gradient descent and show highly encouraging results for splitNN.
\end{abstract}

\section{Introduction}
Collaboration in health is heavily impeded by lack of trust, data sharing regulations such as HIPAA \cite{annas2003hipaa,centers2003hipaa,mercuri2004hipaa,gostin2009beyond,luxton2012mhealth} and limited consent of patients. In settings where different institutions hold different modalities of patient data in the form of electronic health records (EHR), picture archiving and communication systems (PACS) for radiology and other imaging data, pathology test results, or other sensitive data such as genetic markers for disease, collaborative training of distributed machine learning models without any data sharing is desired. Deep learning methods in general have found a pervasive suite of applications in biology, clinical medicine, genomics and public health as surveyed in \cite{ching2018opportunities,shickel2018deep,miotto2017deep,ravi2017deep,alipanahi2015predicting,litjens2017survey}. Training of distributed deep learning models without sharing model architectures and parameters in addition to not sharing raw data is needed to prevent undesirable scrutiny by other entities. As a concrete health example, consider the use case of training a deep learning model for patient diagnosis via collaboration of two entities holding pathology test results and radiology data respectively. These entities are unable to share their raw data with each other due to the concerns noted above. That said, diagnostic performance of the distributed deep learning model is highly contingent on being able to use data from both the institutions for its training. 
In addition to such multi-modal settings, this problem also manifests in settings with entities holding data of the same modality as shown in Fig 1 below. As illustrated, local hospitals or tele-health screening centers do not acquire an enormous number of diagnostic images on their own. These entitites may also be limited by diagnostic manpower. A distributed machine learning method for diagnosis in this setting would enable each individual center to contribute data to an aggregate model without sharing any raw data. This configuration can achieve high accuracy while using significantly lower computational resources and communication bandwidth than previously proposed approaches. This enables smaller hospitals to effectively serve those in need while also benefiting the distributed training network as a whole. In this paper, we build upon splitNN introduced in \cite{gupta2018distributed} to propose specific configurations that cater to practical health settings such as these and furthermore as described in the sections below.

\begin{figure}
    \centering
    \subfigure[Non-cooperating health units]{{\includegraphics[width=6.25cm,height=4.75cm]{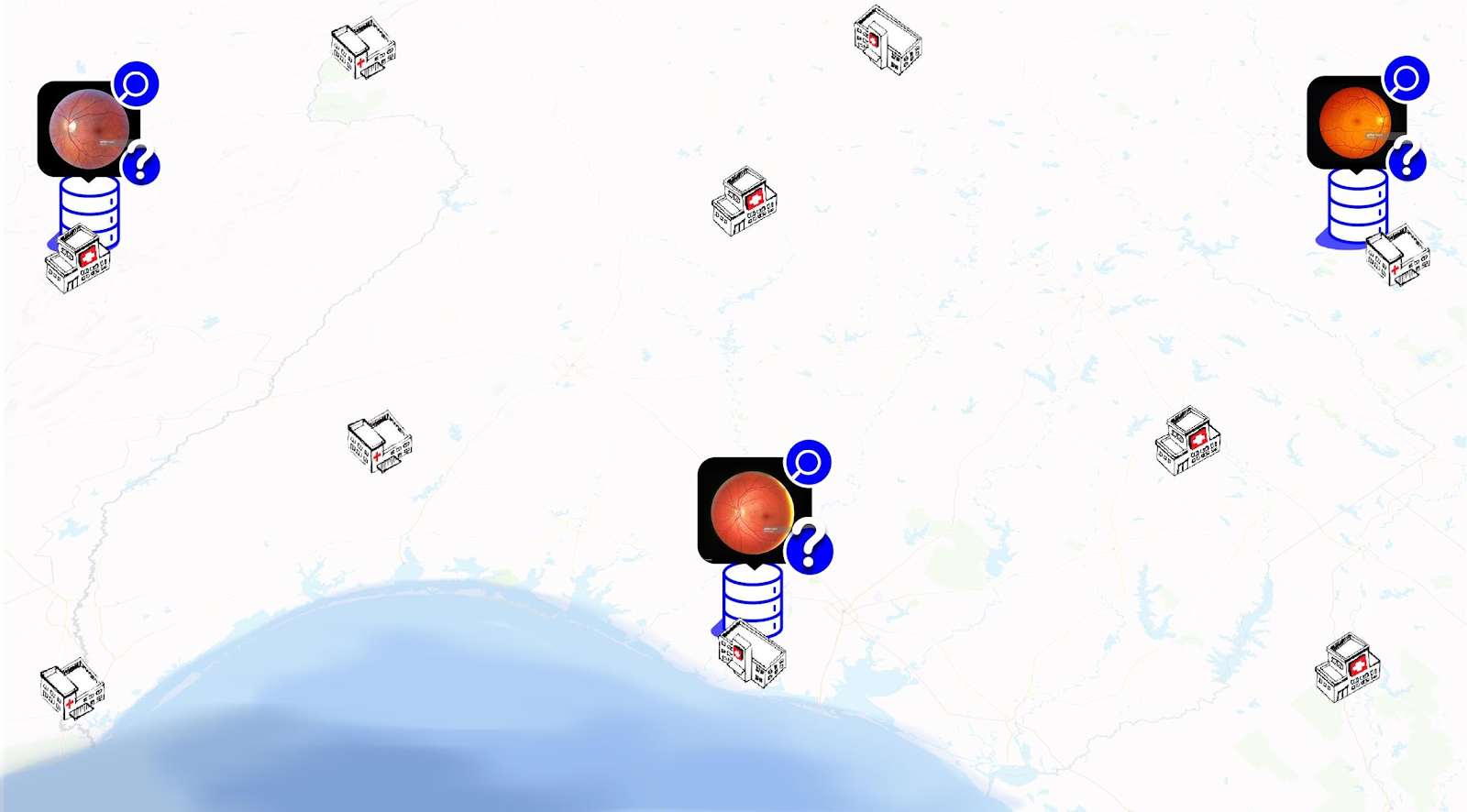} }}
    \qquad
    \subfigure[Distributed learning without raw data sharing
]{{\includegraphics[width=6.25cm,height=4.75cm]{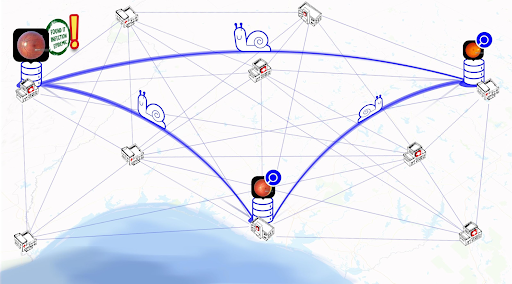} }}%
    \caption{ Distributed learning over retinopathy images (or undetected fast moving threats) over slow bit-rate (‘snail-pace’), to detect the emerging threat by pooling their images but without exchanging raw patient data.
}%
    %\label{fig:example}%
\end{figure} 
%\vspace*{-2mm}
\subsection{Related work:} In addition to splitNN \cite{gupta2018distributed}, techniques of federated deep learning  \cite{mcmahan2016communication} and large batch synchronous stochastic gradient descent (SGD)\cite{chen2016revisiting} are currently available approaches for distributed deep learning. There has been no work as yet on federated deep learning and large batch synchronous SGD methods with regards to their applicability to useful non-vanilla settings of distributed deep learning studied in rest of this paper such as a) distributed deep learning with vertically partitioned data, b) distributed deep learning without label sharing, c) distributed semi-supervised learning and d) distributed multi-task learning. That said, with regards to ‘non-neural network’ based federated learning techniques, the work in \cite{hardy2017private} shows their applicability to vertically partitioned distributed data \cite{navathe1984vertical,agrawal2004integrating,smith2017federated,abadi2007scalable} shows applicability to multi-task learning in distributed settings. We now propose configurations of splitNN for all these useful settings in the rest of this paper. 
 
 \begin{figure}
    \centering
    \subfigure[Simple vanilla split learning]{{\includegraphics[width=4.25cm,height=6cm]{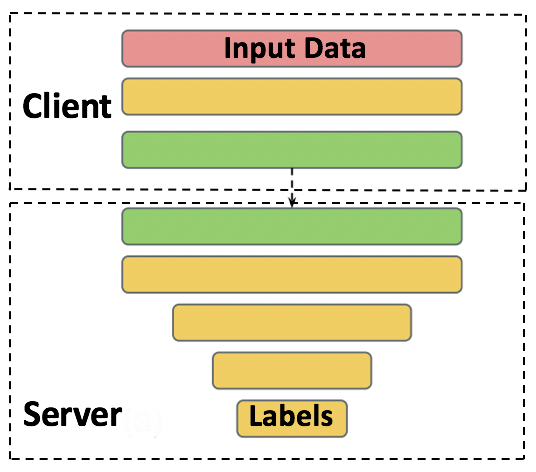} }}
    \quad
    \subfigure[Split learning without label sharing
]{{\includegraphics[width=4.25cm,height=6cm]{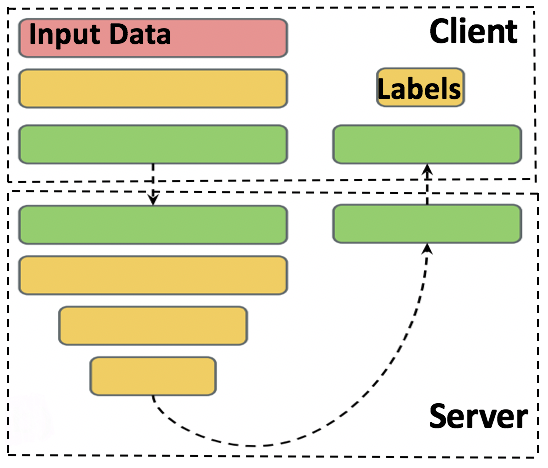} }}%
\quad
 \subfigure[Split learning for vertically partitioned data]{{\includegraphics[width=4.25cm,height=6cm]{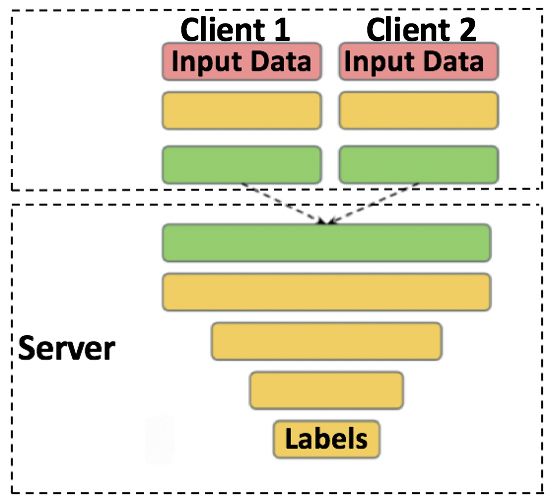} }}
    \caption{Split learning configurations for health shows raw data is not transferred between the client and server health entities for training and inference of distributed deep learning models with SplitNN.   
}%
    %\label{fig:example}%
\end{figure}

\section{SplitNN configurations for health}In this section we propose several configurations of splitNN for various practical health settings:\\\\
\textbf{Simple vanilla configuration for split learning: }This is the simplest of splitNN configurations as shown in Fig 2a. In this setting each client, (for example, radiology center) trains a partial deep network up to a specific layer known as the “cut layer.” The outputs at the cut layer are sent to a server which completes the rest of the training without looking at raw data (radiology images) from clients. This completes a round of forward propagation without sharing raw data. The gradients are now back propagated at the server from its last layer until the cut layer. The gradients at the cut layer (and only these gradients) are sent back to radiology client centers. The rest of back propagation is now completed at the radiology client centers. This process is continued until the distributed split learning network is trained without looking at each others raw data.\\\\
\textbf{U-shaped configurations for split learning without label sharing: }The other two configurations described in this section involve sharing of labels although they do not share any raw input data with each other. We can completely mitigate this problem by a U-shaped configuration that does not require any label sharing by clients. In this setup we wrap the network around at end layers of server’s network and send the outputs back to client entities as seen in Fig.2b. While the server still retains a majority of its layers, the clients generate the gradients from the end layers and use them for backpropagation without sharing the corresponding labels. In cases where labels include highly sensitive information like the disease status of patients, this setup is ideal for distributed deep learning.
\\\\
\textbf{Vertically partitioned data for split learning: }This configuration allows for multiple institutions holding different modalities of patient data  \cite{hardy2017private,navathe1984vertical,agrawal2004integrating} to learn distributed models without data sharing. In Fig. 2c, we show an example configurations of splitNN suitable for such multi-modal multi-institutional collaboration. As a concrete example we walkthrough the case where radiology centers collaborate with pathology test centers and a server for disease diagnosis. As shown in Fig. 2c radiology centers holding imaging data modalities train a partial model upto the cut layer. In the same way the pathology test center having patient test results trains a partial model upto its own cut layer. The outputs at the cut layer from both these centers are then concatenated and sent to the disease diagnosis server that trains the rest of the model. This process is continued back and forth to complete the forward and backward propagations in order to train the distributed deep learning model without sharing each others raw data. 
We would like to note that although these example configurations show some versatile applications for splitNN, they are by no means the only possible configurations. 
\section{Results about resource efficiency
}We share a comparison from  \cite{gupta2018distributed} of validation accuracy and required client computational resources in Figure 3 for the three techniques of federated learning, large batch synchronous SGD and splitNN as they are tailored for distributed deep learning. 
\begin{figure}
    \centering
    \subfigure[Accuracy vs client-side flops on 100 clients with
VGG on CIFAR 10]{{\includegraphics[width=6.5cm,height=7cm]{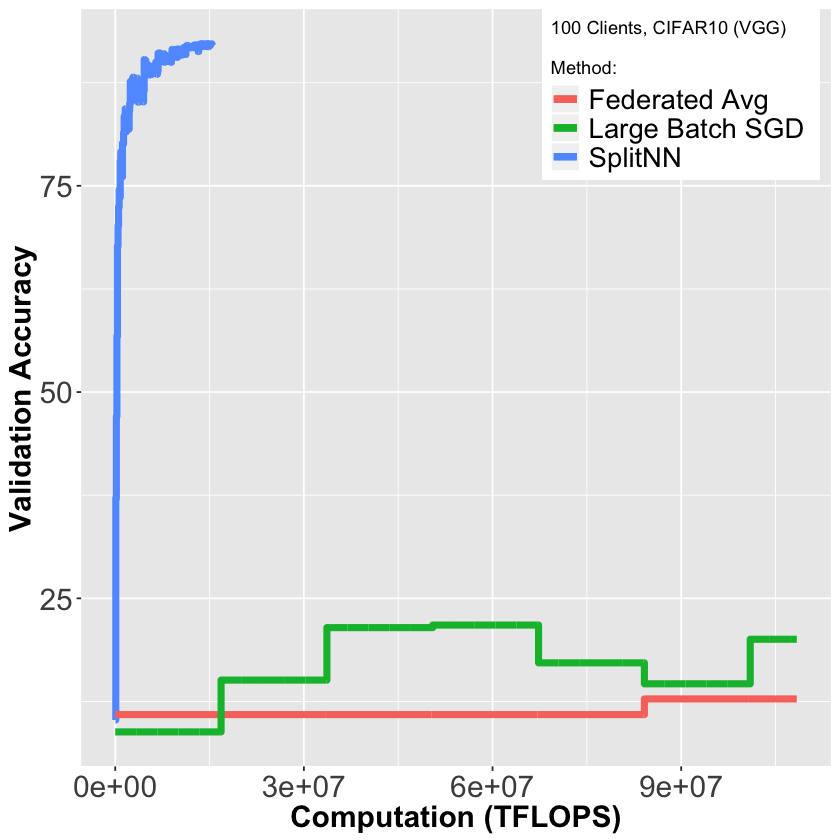} }}
    \qquad
    \subfigure[Accuracy vs client-side flops on 500 clients with
Resnet-50 on CIFAR 100
]{{\includegraphics[width=6.5cm,height=7cm]{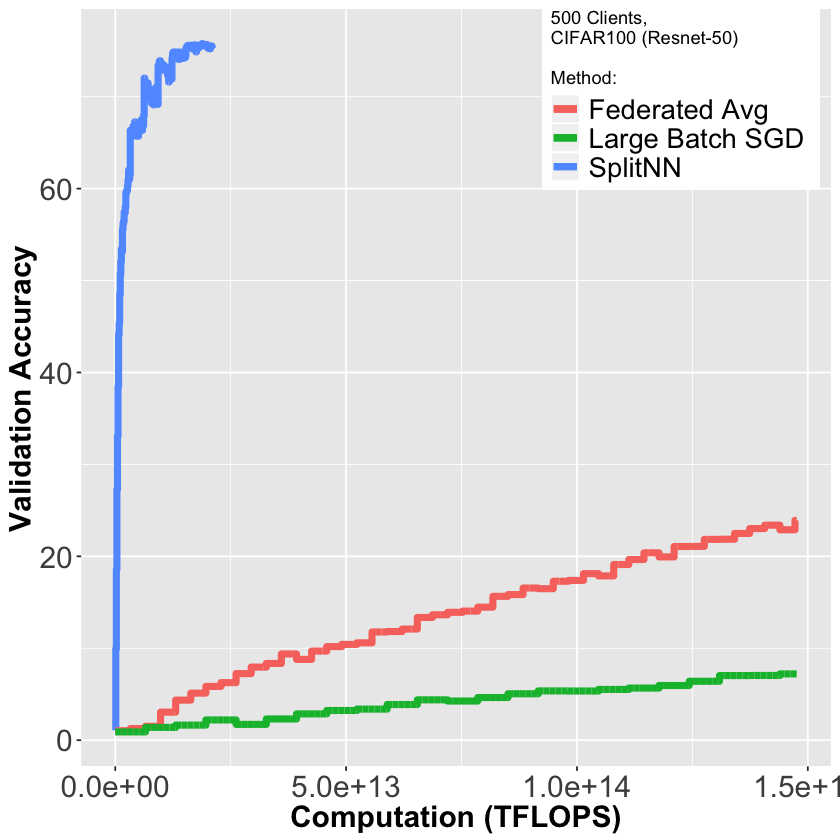} }}%
    \caption{We show dramatic reduction in computational burden (in tflops) while maintaining higher accuracies when training over large number of clients with splitNN. Blue line denotes distributed deep learning using splitNN, red line indicate federated averaging and green line indicates large batch SGD.}%
    \end{figure}
As seen in this figure, the comparisons were done on the CIFAR 10 and CIFAR 100 datasets using VGG and Resnet-50 architectures for 100 and 500 client based setups respectively. In this distributed learning experiment we clearly see that SplitNN outperforms the techniques of federated learning and large batch synchronous SGD in terms of higher accuracies with drastically lower computational requirements on the side of clients. In tables 1 and 2 we share more comparisons from  \cite{gupta2018distributed} on computing resources in TFlops and communication bandwidth in GB required by these techniques. SplitNN again has a drastic improvement of computational resource efficiency on the client side. In the case with a relatively smaller number of clients the communication bandwidth required by federated learning is less than splitNN. These improvements on the client side resource efficiency are even more dramatic due to the presence of a smaller number of parameters in earlier layers of convolutional neural networks (CNN’s) like VGG and Resnet in addition to the fact that computation is split due to the cut layers. This uneven distribution of network parameters holds for the vast majority of modern CNN’s, a property that SplitNN can effectively exploit. 

\begin{table}[htbp]
\centering
\begin{tabular}{|l|l|l|}
\hline
\textbf{Method}    & \textbf{100 Clients} & \textbf{500 Clients} \\ \hline
\textbf{Large Batch SGD}    & 29.4 TFlops          & 5.89 TFlops          \\ \hline
\textbf{Federated Learning} & 29.4 TFlops          & 5.89 TFlops          \\ \hline
\textbf{SplitNN}            & 0.1548 TFlops        & 0.03 TFlops          \\ \hline
\end{tabular}
\vspace{1em}
\caption{Computation resources consumed per client when training CIFAR 10 over VGG (in teraflops) are drastically lower for SplitNN than Large Batch SGD and Federated Learning.}
\vspace{-1.8em}
\end{table}

\begin{table}[!htbp]
\centering
\begin{tabular}{|l|l|l|}
\hline
\textbf{Method}    & \textbf{100 Clients} & \textbf{500 Clients} \\ \hline
\textbf{Large Batch SGD}    & 13 GB                & 14 GB                \\ \hline
\textbf{Federated Learning} & 3 GB                 & 2.4 GB               \\ \hline
\textbf{SplitNN}            & 6 GB                 & 1.2 GB               \\ \hline
\end{tabular}
\vspace{1em}
\caption{Computation bandwidth required per client when training CIFAR 100 over ResNet (in gigabytes) is lower for splitNN than large batch SGD and federated learning with a large number of clients. For setups with a smaller number of clients, federated learning requires a lower bandwidth than splitNN. Large batch SGD methods popular in data centers use a heavy bandwidth in both settings.}
\end{table}
% \vspace*{-5mm}
\section{Conclusion and future work}
Simple configurations of distributed deep learning do not suffice for various practical setups of collaboration across health entities. We propose novel configurations of a recently proposed distributed deep learning technique called splitNN that is dramatically resource efficient in comparison to currently available distributed deep learning methods of federated learning and large batch synchronous SGD. SplitNN is versatile in allowing for many plug and play configurations based on the required application. Generaton of such novel configurations in health and beyond is a good avenue for future work. SplitNN is also scalable to large-scale settings and can use any state of the art deep learning architectures. In addition, the boundaries of resource efficiency can be pushed further in distributed deep learning by combining splitNN with neural network compression methods \cite{lin2017deep,louizos2017bayesian,han2015deep} for seamless distributed learning with edge devices.

   \section{Supplementary material:}
  \subsection{Additional configurations}
  In this supplementary section we propose some more split learning configurations of splitNN for versatile collaborations in health to train and infer from distributed deep learning models without sharing raw patient data.

\textbf{Extended vanilla split learning:}
As shown in Fig. 4a we give another modification of vanilla split learning where the result of concatenated outputs is further processed at another client before passing it to the server.\\\\
\textbf{Configurations for multi-task split learning:}
As shown in Fig. 4b, in this configuration multi-modal data from different clients is used to train partial networks up to their corresponding cut layers. The outputs from each of these cut layers are concatenated and then sent over to multiple servers. These are used by each server to train multiple models that solve different supervised learning tasks.\\\\
\textbf{Tor \cite{syverson2004tor} like configuration for multi-hop split learning: }This configuration is an analogous extension of the vanilla configuration. In this setting multiple clients train partial networks in sequence where each client trains up to a cut layer and transfers its outputs to the next client. This process is continued as shown in Fig. 4c as the final client sends its activations from its cut layer to a server to complete the training.\\
 
 \begin{figure*}[htbp]
    \centering
    \subfigure[Extended vanilla split learning]{{\includegraphics[width=4.25cm,height=5cm]{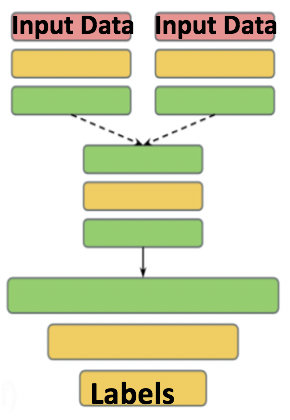} }}
    \quad
    \subfigure[Split learning for multi-task output with vertically partitioned input]{{\includegraphics[width=4.25cm,height=5cm]{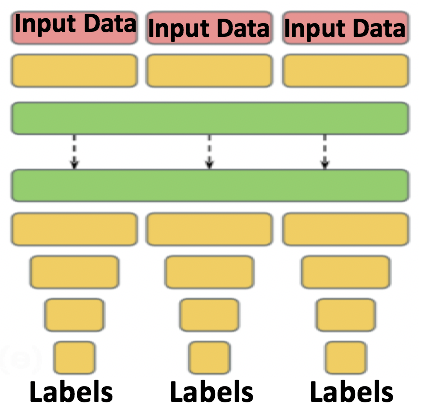} }}%
\quad
 \subfigure['Tor'\cite{syverson2004tor} like multi-hop split learning]{{\includegraphics[width=4.25cm,height=5cm]{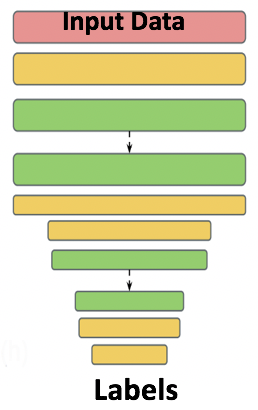} }}
    \caption{Split learning configurations for health shows raw data is not transferred between the client and server health entities for training and inference of distributed deep learning models with SplitNN.   
}%
   % \label{fig:example}%
\end{figure*} 
  
 % \printbibliography

 \end{document}